\documentclass{article}

\usepackage[final]{neurips_2020}

\usepackage[utf8]{inputenc} 
\usepackage[T1]{fontenc}    
\usepackage{hyperref}       
\usepackage{url}            
\usepackage{booktabs}       
\usepackage{amsfonts}       
\usepackage{nicefrac}       
\usepackage{microtype}      
\usepackage{subcaption}
\usepackage{graphicx}
\usepackage{booktabs}

\title{MonarchNet: Differentiating Monarch Butterflies from Butterflies Species with Similar Phenotypes}

\author{%
  Thomas Y. ~Chen\\
  The Academy for Mathematics, Science, and Engineering\\
  \texttt{thomaschen7@acm.org}\\
}

\begin{document}

\maketitle

\begin{abstract}
In recent years, the monarch butterfly's iconic migration patterns have come under threat from a number of factors, from climate change to pesticide use. To track trends in their populations, scientists as well as citizen scientists must identify individuals accurately. This is uniquely key for the study of monarch butterflies because there exist other species of butterfly, such as viceroy butterflies, that are "look-alikes" (coined by the Convention on International Trade in Endangered Species of Wild Fauna and Flora), having similar phenotypes. To tackle this problem and to aid in more efficient identification, we present MonarchNet, the first comprehensive dataset consisting of butterfly imagery for monarchs and five look-alike species. We train a baseline deep-learning classification model to serve as a tool for differentiating monarch butterflies and its various look-alikes. We seek to contribute to the study of biodiversity and butterfly ecology by providing a novel method for computational classification of these particular butterfly species. The ultimate aim is to help scientists track monarch butterfly population and migration trends in the most precise and efficient manner possible.
  
\end{abstract}
\section{Introduction}
The famous monarch butterflies (\textit{Danaus plexippus}) and their migration patterns are an iconic phenomenon of the natural world. The species' unique behavior has warranted a variety of in-depth studies over the years \cite{overwintering1989reproductive, wassenaar1998natal, mouritsen2002virtual, reppert2010navigational, heinze2011sun, kass2020biotic}. Monarch butterflies are important pollinators and crucial members of the ecosystems to which they contribute \cite{landis2014create}. However, in recent times, there have been concerns about dramatic declines in population and disruptions to their migrations, caused by a variety of reasons, including climate change, deforestation, droughts, loss of habitat, and loss of milkweed \cite{brower2012decline, agrawal2019advances, thogmartin2017monarch}. Notably, citizen science has been crucial to discovering and monitoring these troubling declines \cite{schultz2017citizen}. In order to gather data and gain important insights into monarch butterfly trends, scientists and citizen scientists must first identify the insects accurately and efficiently. 

\begin{figure}[h!]
  \centering
  \begin{subfigure}[b]{0.4\linewidth}
    \includegraphics[width=\linewidth]{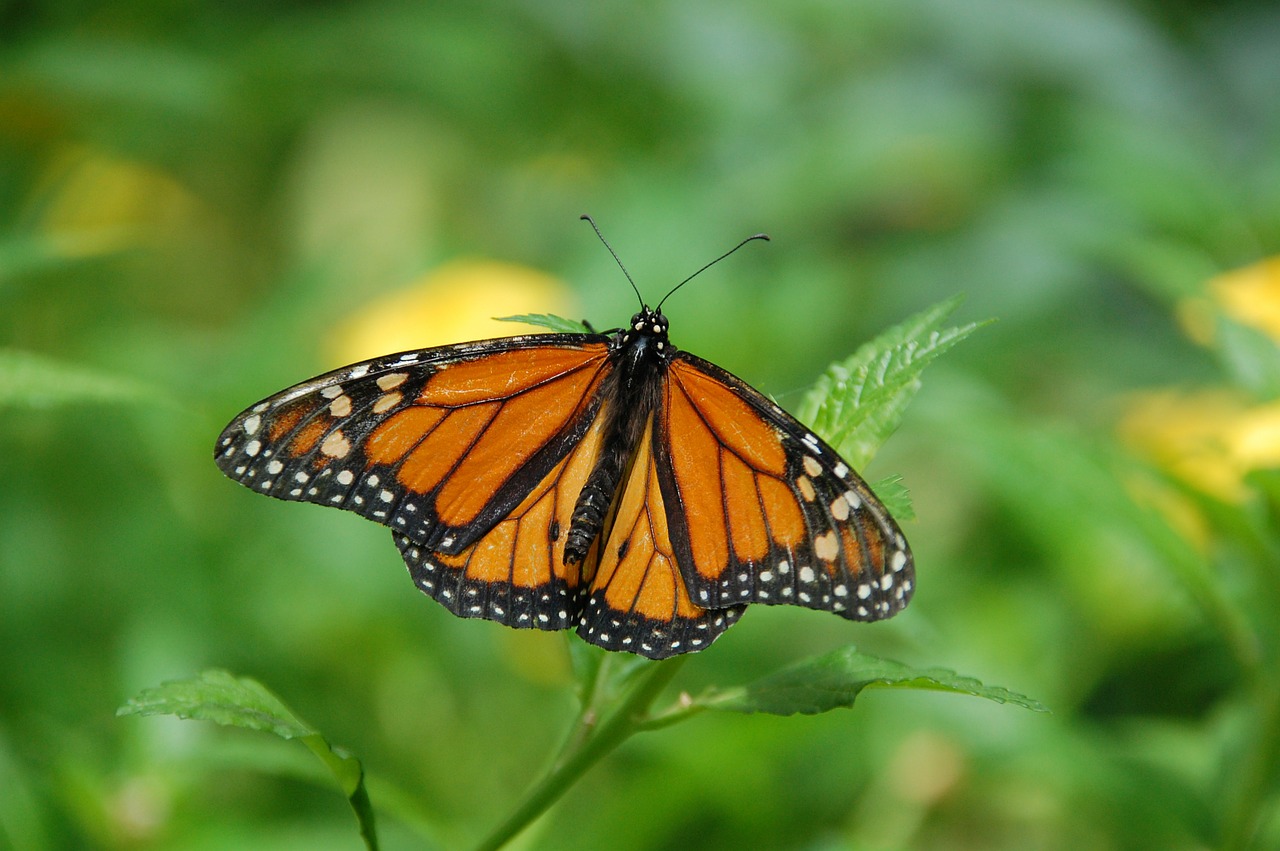}
    \caption{Monarch Butterfly (\textit{Danaus plexippus})}
  \end{subfigure}
  \begin{subfigure}[b]{0.4\linewidth}
    \includegraphics[width=\linewidth]{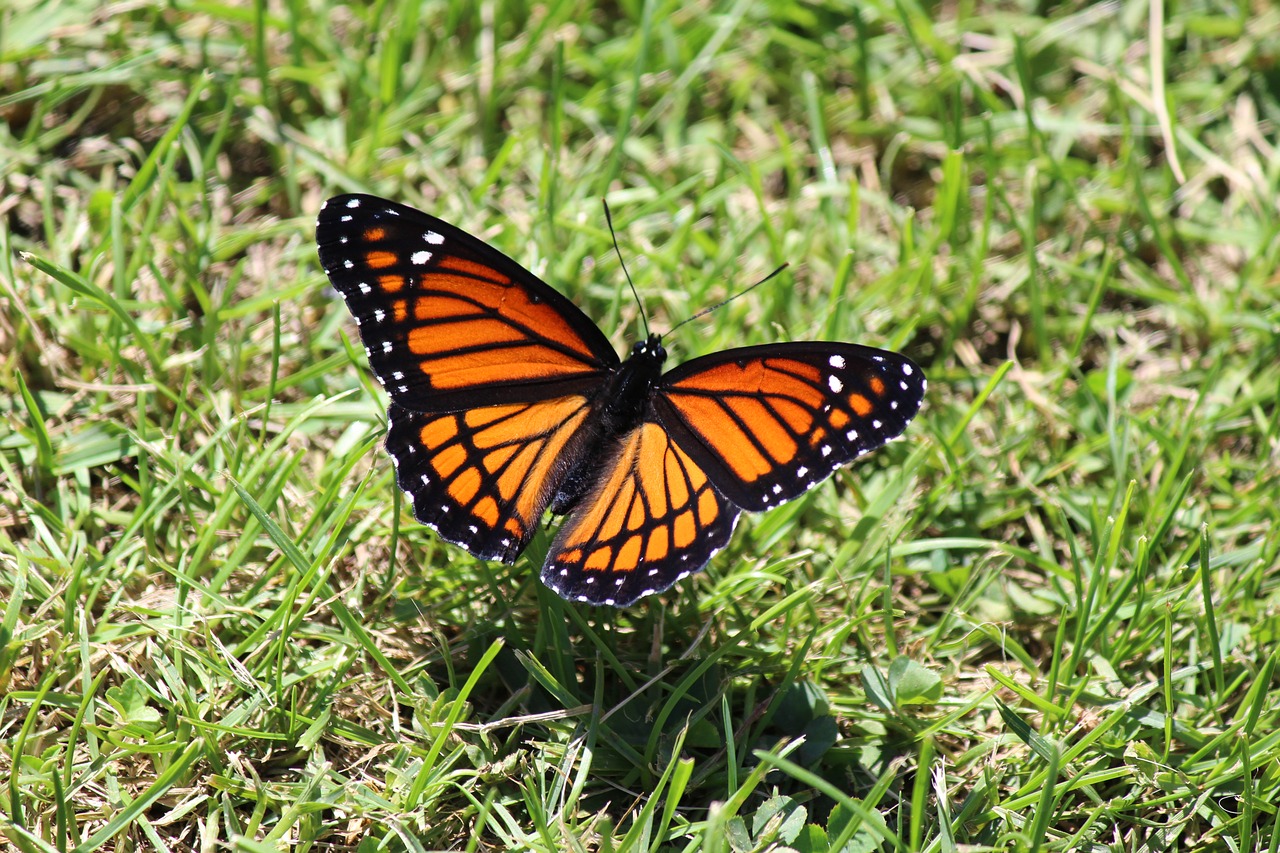}
    \caption{Viceroy Butterfly (\textit{Limenitis archippus})}
  \end{subfigure}
  \begin{subfigure}[b]{0.4\linewidth}
    \includegraphics[width=\linewidth]{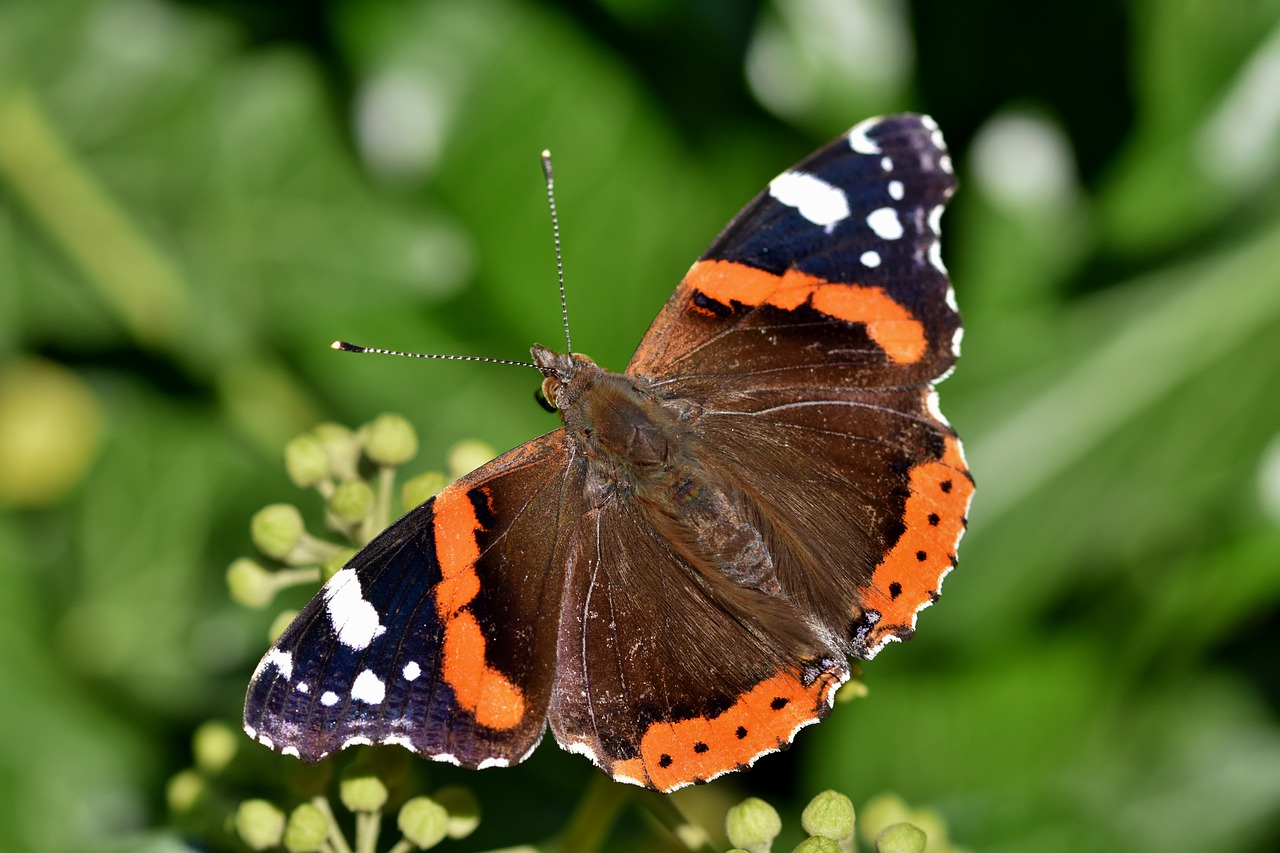}
    \caption{Red Admiral (\textit{Vanessa atalanta})}
  \end{subfigure}
  \begin{subfigure}[b]{0.4\linewidth}
    \includegraphics[width=\linewidth]{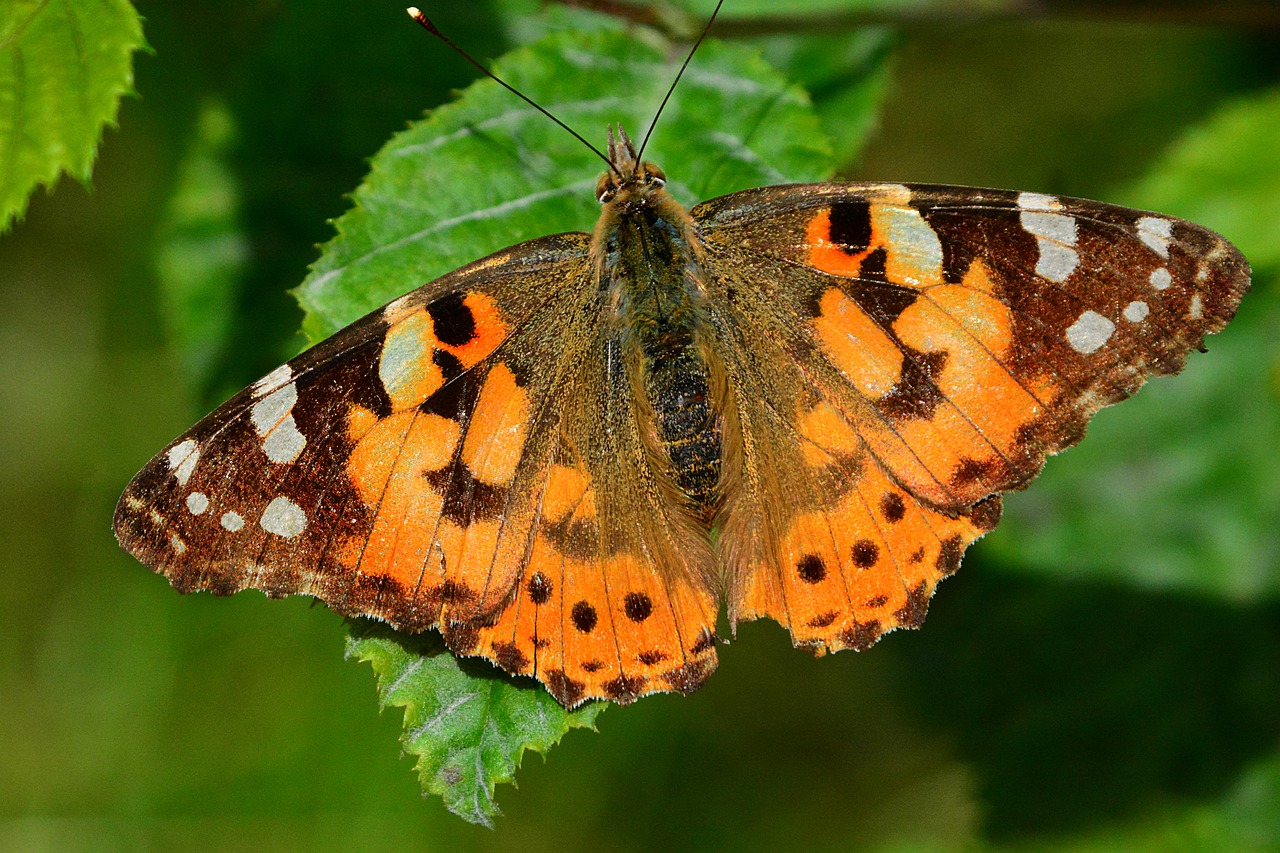}
    \caption{Painted Lady (\textit{Vanessa cardui})}
  \end{subfigure}
  \begin{subfigure}[b]{0.4\linewidth}
    \includegraphics[width=\linewidth]{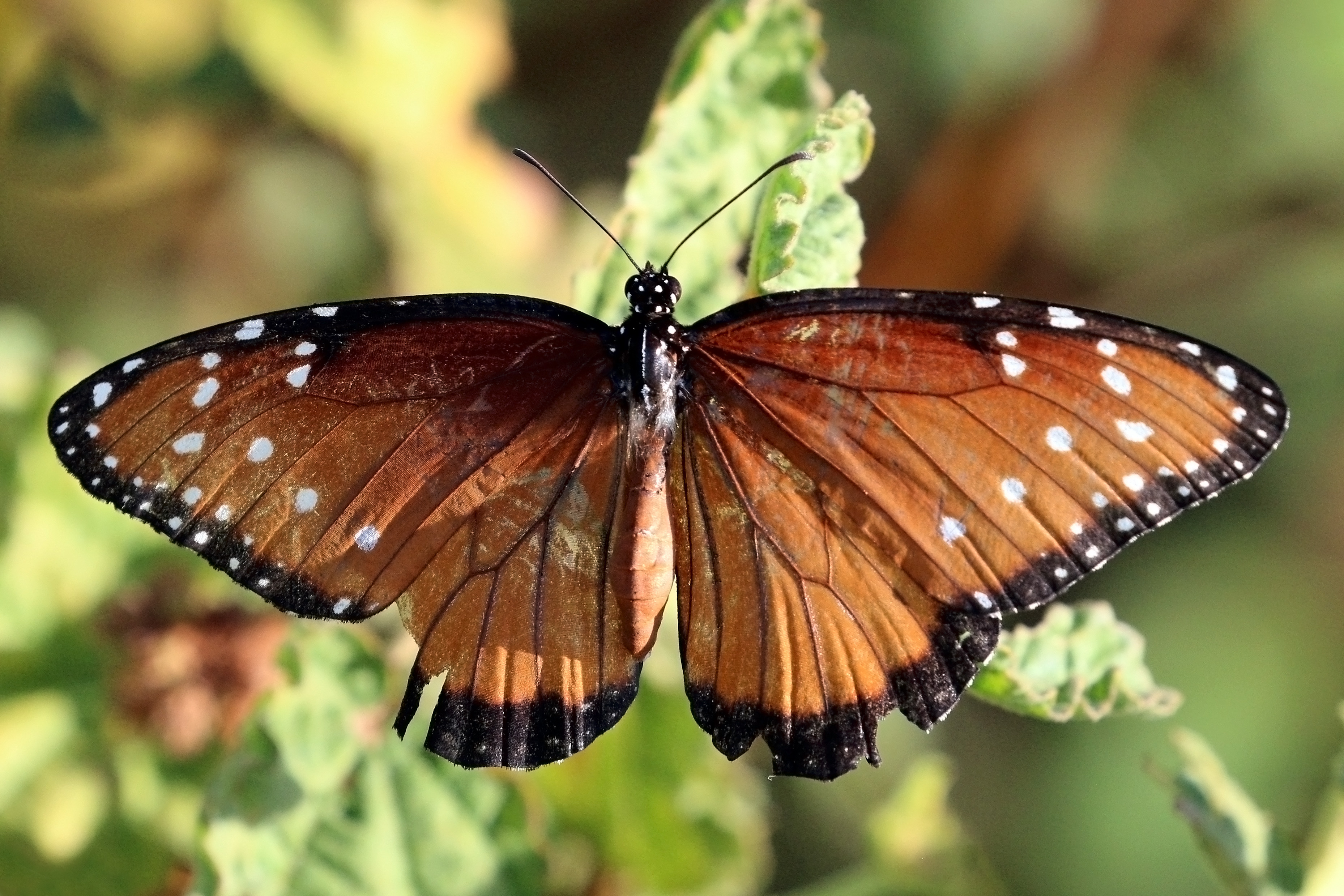}
    \caption{Queen Butterfly (\textit{Danaus gilippus})}
  \end{subfigure}
  \begin{subfigure}[b]{0.4\linewidth}
    \includegraphics[width=\linewidth]{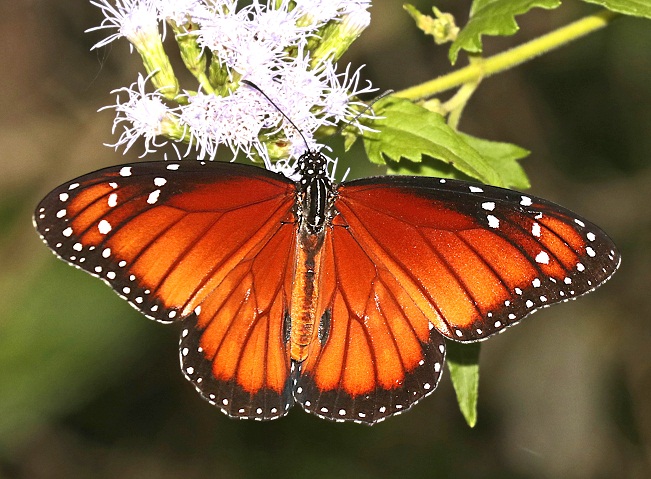}
    \caption{Soldier Butterfly (\textit{Danaus eresimus})}
  \end{subfigure}
  \caption{Sample images of a monarch butterfly and members of 5 phenotypically similar species}
  \label{fig:butterflies}
\end{figure}

Monarch butterflies have many counterparts that have similar external appearances, from viceroy butterflies (\textit{Limenitis archippus}) to soldier butterflies (\textit{Danaus eresimus}). The five species displayed in Figure \ref{fig:butterflies} are of particular interest. In some cases, particularly for the viceroy butterfly, these similarities are due to mimicry (specifically Müllerian mimicry) \cite{alexander2019mechanism}. In this case, the main visual difference is the black line across the viceroy's hind wings. Hence, recognizing monarch butterflies correctly and not unknowingly observing an entirely different species by mistake can be a challenge when studying the species. To approach this problem with machine learning, we must train deep neural networks on butterfly image data to enable instantaneous and accurate classification.

Currently, there are not any image datasets that incorporate both monarch butterflies and their close neighbors in physical appearance. Thus, in this paper, we present a novel butterfly imagery dataset, MonarchNet, for the purpose of classifying butterflies that have a similar physical phenotype. We also train a baseline convolutional neural network (CNN) on this dataset for assessment.

\section{Dataset Overview}
MonarchNet contains labeled images of monarch butterflies (\textit{Danaus plexippus}) and of species that have a similar physical appearance: viceroy butterflies (\textit{Limenitis archippus}), red admirals (\textit{Vanessa atalanta}), painted ladies (\textit{Vanessa cardui}), queen butterflies (\textit{Danaus gilippus}), and soldier butterflies (\textit{Danaus eresimus}). These categories contain 200192, 98765, 53654, 26578, 12534, and 14213 images, respectively, with each image containing exactly one butterfly.

\section{Dataset Curation}
\begin{figure}[h!]
  \centering
  \includegraphics[width=100mm]{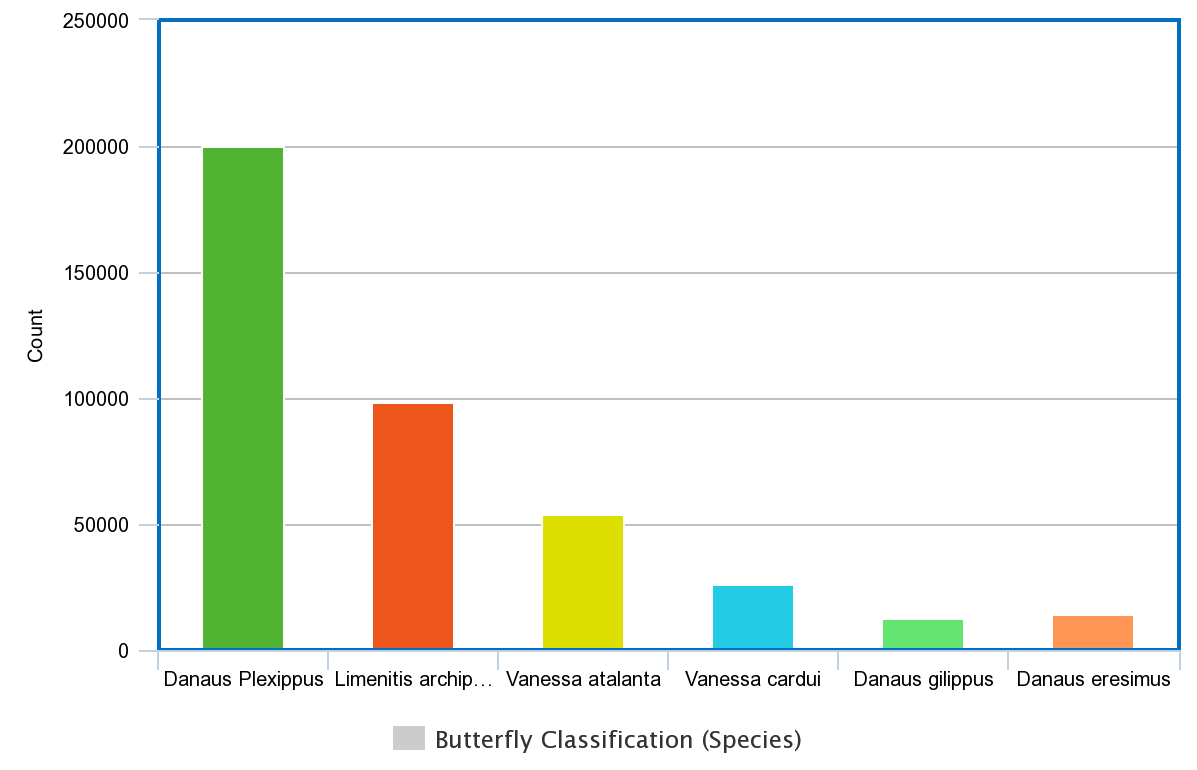}
  \caption{Distribution of Butterfly Species Labels}
  \label{dist}
\end{figure}

Image data for the MonarchNet dataset are sourced and downloaded from Google Images. This was chosen because it was determined to be the most optimal way to feasibly gather a large and sufficient volume of data for each of the six butterfly species categories. Only images with the Creative Commons Licenses were used and image urls are saved. Queries of both species' common and scientific name (i.e. viceroy butterfly as well as \textit{Limenitis archippus}) were used to obtain images. For all queries (scientific or colloquial), we label the downloaded images with the relevant scientific name. Duplicate or near-duplicate images are discarded using image hashing. We obtain 405,936 images in total, with a distribution displayed in Figure \ref{dist}. Additionally, bounding boxes for the butterflies are annotated and provided as labels.

However, images sourced from Google Images are noisy and must be filtered. We manually clean the data and verify that images fit their respective labels. This is done through re-examining the image source or verifying specific characteristics relevant to the corresponding species.

\section{Baseline Model}
We train a baseline CNN to evaluate the dataset and its usefulness for the monarch butterfly versus "look-alike" species classification problem. The input consists of a labeled butterfly image, with the species and bounding box coordinates. The output is a digit from 0 to 5 representing the predicted species of the butterfly in the image. 0, 1, 2, 3, 4, and 5 correspond to the classes a, b, c, d, e, and f shown in Figure \ref{fig:butterflies}, respectively. The model architecture is ResNet50 \cite{he2016deep}, pretrained on ImageNet data \cite{deng2009imagenet}. The criterion for optimization is the cross-entropy loss function. We train on a randomly selected 80\% of the dataset with a batch size of 32. The Adam optimizer with a learning rate of 0.01 is utilized. The network is trained for 100 epochs on NVIDIA Tesla K80 GPUs.
\subsection{Results and Discussion}
Testing on the remaining 20\% of the data, our baseline model achieves a weighted F1 score of 0.824. F1 scores for individual categories are displayed in Table \ref{tab:f1}. This measure of evaluation is used due to the imbalanced distribution of the dataset. For example, if simple accuracy was our metric, if the model always predicted the monarch butterfly (\textit{Danaus plexippus}) category for each image input, we would see an artificially inflated percentage because images with that particular label make up a disproportionately large part of the dataset (Fig. \ref{dist}).
\begin{table}[h!]
\begin{center}
\caption{Baseline Model F1 Scores}
\label{tab:f1}

\begin{tabular}{@{}ll@{}}
\toprule
Butterfly Species            & F1 Score \\ \midrule
\textit{Danaus plexippus} (0)    & 0.795         \\
\textit{Limenitis archippus} (1) & 0.761         \\
\textit{Vanessa atalanta} (2)    & 0.962         \\
\textit{Vanessa cardui} (3)      & 0.975         \\
\textit{Danaus gilippus} (4)     & 0.823         \\
\textit{Danaus eresimus} (5)     & 0.854         \\ \bottomrule
\end{tabular}%
\end{center}
\end{table}

From Table \ref{tab:f1}, we see that the model performs best, as per F1 score, on \textit{Vanessa atalanta} and \textit{Vanessa cardui}. This is a reasonable result because these two species are more physically differentiable, for both humans and AI, from the other species included in MonarchNet. A brief glance at Figure \ref{fig:butterflies} reveals relatively significant phenotypical differences. The CNN performs the worst on images with the ground truth label \textit{Danaus plexippus} and \textit{Limenitis archippus}. This is also justifiable because the monarch and viceroy butterflies are arguably the most similar in characteristics and may cause confusion. The remaining two species lie in the middle in terms of the model's performance.

\section{Conclusion}
In this work, we presented the MonarchNet dataset and a corresponding baseline CNN to provide a machine-learning-based approach to visually differentiating between monarch butterflies and five other species that appear similarly. This research contributes to monarch butterfly research as well as the study of the biodiversity and ecology of relevant ecosystems by introducing a novel method for tackling this problem. This is crucial to the monitoring of the decline of monarch butterflies and the assessment and mitigation of threats to populations. Our AI method is especially helpful for citizen scientists, who are vital to these efforts. Effective conservation strategies and habitat restoration efforts can only be developed if accurate data is gathered. Future work may include curating datasets that incorporate a larger range of butterfly species, in order to aid in conservation efforts around the globe, not solely limited to monarchs (for example, in the Amazon rainforest). The computer vision community is encouraged to strive to improve upon the baseline model presented here when the dataset and code are released.
\bibliography{mybib}{}
\bibliographystyle{plain}

\end{document}